\documentclass{article} % For LaTeX2e
\usepackage{iclr2024_conference,times}
\usepackage{amsmath,amsfonts,bm}

\def\eqref#1{equation~\ref{#1}}
\def\1{\bm{1}}

\DeclareMathAlphabet{\mathsfit}{\encodingdefault}{\sfdefault}{m}{sl}
\SetMathAlphabet{\mathsfit}{bold}{\encodingdefault}{\sfdefault}{bx}{n}

\usepackage{hyperref}
\usepackage{url}
\usepackage{graphicx}
\usepackage{booktabs}
\usepackage{adjustbox}
\usepackage{tabularx}
\usepackage{subcaption}

\title{Temporal Cross-Attention for Dynamic \\ Embedding and Tokenization of Multimodal Electronic Health Records}
\author{Yingbo Ma, Suraj Kolla, Dhruv Kaliraman, Victoria Nolan, Zhenhong Hu, \\ \textbf{Ziyuan Guan, Yuanfang Ren, Brooke Armfield, Tezcan Ozrazgat-Baslanti} \\
Department of Medicine, University of Florida\\
\texttt{\{yingbo.ma,n.kolla,dhruv.kaliram,vnolan,hzhuf\}@ufl.edu} \\ 
\texttt{\{ziyuan.guan,renyuanfang,barmfield,tezcan\}@ufl.edu}
\\
\And
Tyler J. Loftus \\
Department of Surgery, University of Florida\\
\texttt{tloftus@ufl.edu} \\
\AND
Parisa Rashidi \\
Department of Biomedical Engineering, University of Florida\\
\texttt{parisa.rashidi@ufl.edu} \\
\AND
Azra Bihorac$^{*}$, Benjamin Shickel\thanks{Authors contributed equally}\\
Department of Medicine, University of Florida\\
\texttt{\{abihorac,shickelb\}@ufl.edu} \\
}

\iclrfinalcopy

\begin{document}

\maketitle

\begin{abstract}
The breadth, scale, and temporal granularity of modern electronic health records (EHR) systems offers great potential for estimating personalized and contextual patient health trajectories using sequential deep learning. However, learning useful representations of EHR data is challenging due to its high dimensionality, sparsity, multimodality, irregular and variable-specific recording frequency, and timestamp duplication when multiple measurements are recorded simultaneously. Although recent efforts to fuse structured EHR and unstructured clinical notes suggest the potential for more accurate prediction of clinical outcomes, less focus has been placed on EHR embedding approaches that directly address temporal EHR challenges by learning time-aware representations from multimodal patient time series. In this paper, we introduce a dynamic embedding and tokenization framework for precise representation of multimodal clinical time series that combines novel methods for encoding time and sequential position with temporal cross-attention. Our embedding and tokenization framework, when integrated into a multitask transformer classifier with sliding window attention, outperformed baseline approaches on the exemplar task of predicting the occurrence of nine postoperative complications of more than 120,000 major inpatient surgeries using multimodal data from three hospitals and two academic health centers in the United States. 
\end{abstract}

\section{Introduction}
Electronic health records (EHRs) contain important information about patient encounters that support real-world healthcare delivery, and while artificial intelligence and machine learning have the theoretical potential to support clinical decision-making based on contextual representations of patient data, modeling real-world EHR time series is challenging due to its high dimensionality, multimodality, and temporal characteristics of the healthcare domain.

While evidence suggests that sequential deep learning approaches can outperform conventional machine learning for patient-level predictions \citep{shickel2023dynamic,adiyeke2023deep,morid2023time}, popular approaches such as recurrent neural networks (RNN) with long short-term memory (LSTM) \citep{memory2010long} and gated recurrent networks \citep{chung2014empirical} do not account for the temporal complexities of EHR data and may be suboptimal when learning temporal dynamics of patient health trajectories. Recently, transformers have been used for modeling temporal EHR data \citep{li2020behrt,shickel2022multi,tipirneni2022self} and have been established as state-of-the-art approaches for predicting clinical outcomes from patient data sequences.

However, additional challenges persist when modeling EHR data with transformers, such as capturing temporal dependency across very long sequences \citep{li2022hi} and modeling heterogeneous dependencies across variables \citep{zhang2022crossformer}. Unstructured clinical notes, which contain important information about a patient encounter \citep{jensen2017analysis}, have the potential to provide added context to structured EHR. Recent studies have shown performance improvements obtained from jointly modeling structured EHR data and unstructured clinical notes for various multimodal clinical prediction tasks \citep{liu2022dynamic,zhang2023improving}. However, how to effectively learn the multimodal EHR representations in light of temporal EHR complexities remains an open question.

In this paper, we introduce a dynamic embedding and tokenization scheme to enable transformers to adapt to the unique challenges of multimodal clinical time series. Our scheme uses flexible positional encoding and a learnable time embedding to address the challenge of sparsity and irregular sampling, and a variable-specific encoding strategy for capturing distinct characteristics and relationships between temporal variables. To effectively combine structured, numerical time series data and free-text clinical notes, we adopted a cross-attention-based approach that learns a joint multimodal temporal representation. We demonstrate the effectiveness of our approach and analyze the relative contributions of each component of our framework using the benchmark task of predicting the onset of multiple postoperative complications following major inpatient surgery.

%The task in this paper can be setup as this: considering a dataset $\mathbf{D}=\left ( P_i, \mathbf{d}_i, \mathbf{T}_i, \mathbf{N}_i, \mathbf{y}_i \right )$, in which for each patient $P_{i}$, the dataset contains the patient's demographic information $\mathbf{d}_i$, a multivariate time series $\mathbf{T}_i$, corresponding clinical notes $\mathbf{N}_i$, and clinical outcomes $\mathbf{y}_i$. In this work, the task objective is to predict $\mathbf{y}_i$, given $\left (\mathbf{d}_i, \mathbf{T}_i, \mathbf{N}_i\right )$.

\section{Methods}
Our dynamic embedding and tokenization framework for multimodal clinical time series includes methods designed to address (a) the temporal complexities of real-world EHR, and (b) the multimodal integration of structured EHR and unstructured clinical notes.

\subsection{Encoding dynamic temporality in clinical time series}
\begin{figure}[h]
\centering
\includegraphics[width=.9\textwidth]{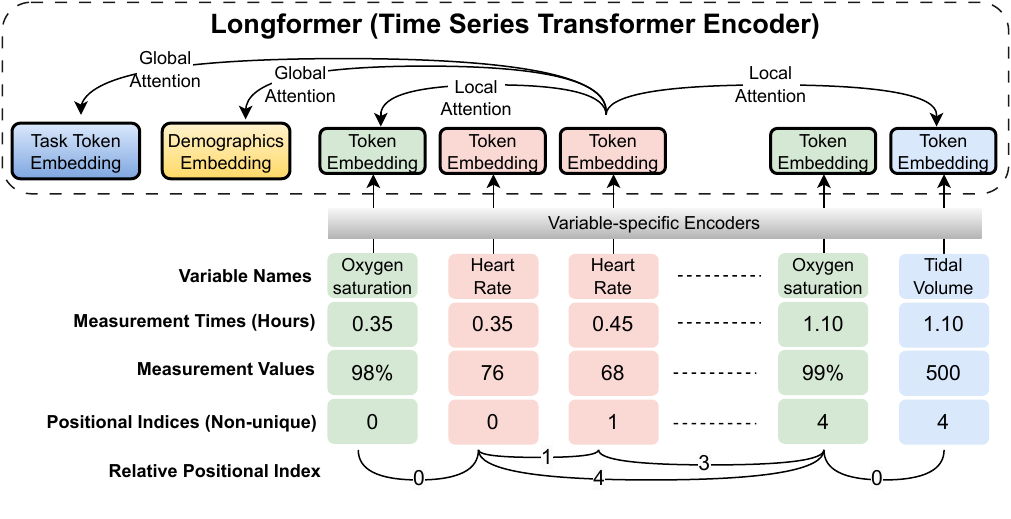}
\caption{Dynamic embedding and tokenization scheme for multivariate clinical time series.}
\label{fig:1}
\end{figure}

Figure~\ref{fig:1} shows an overview of our dynamic embedding and tokenization scheme, which introduces three novelties to existing approaches: a flexible positional encoding, a learnable time encoding, and variable-specific encoding.

\textbf{Flexible positional encoding}. Multivariate EHR time series contain variables measured at different frequencies (e.g., a vital sign saved every minute vs. a laboratory test taken every 24 hours), and other variables that are measured at exactly the same time (e.g., blood pressure, heart rate, and respiratory rate from a bedside monitor). Traditional approaches that enforce a single resampled time interval, or present duplicate-time inputs in an arbitrary ordering, have the potential to lose information or inject unnecessary bias into the data representation. Our approach uses non-unique absolute positional indices based on the recorded timestamps so that variable tokens measured at the same time will be assigned the same positional index. To model the relationships between clusters of short-term activity across a long timeframe, we add a relative positional encoding to each token embedding \citep{shaw2018self}, which can help capture local token dependencies, especially for processing long sequences\citep{zaheer2020big,wei2021position}.

\textbf{Learnable time encoding}. Positional embeddings are used by models such as transformers to inject information about sequential order into time series representations \citep{dufter2022position}. However, positional embeddings alone omit critical information about the relative time between events. For applications of transformers to time series, time embeddings can help capture important temporal patterns \citep{zhang2020time,zeng2023transformers}. Our dynamic tokenization scheme uses Time2Vec \citep{kazemi2019time2vec} to learn a model-agnostic vector representation for time. In Time2Vec, a time \textit{t} is encoded to generate one non-periodic $\omega_{np}  t + \phi_{np} $, and one periodic $sin(\omega_{p}  t + \phi_{p})$ time dependent vector, where $\omega$ and $\phi$ are learnable parameters \citep{liang2023learn}.

\textbf{Variable-specific encoding}. A multivariate clinical time series often includes different categories of health variables (e.g., vital signs, laboratory tests, medications) that tend to exhibit distinct characteristics, numerical ranges, and temporal patterns. In recent EHR transformer implementations like BEHRT \citep{li2020behrt}), all tokens within the sequence are embedded with the same encoder, which does not account for the heterogeneity among different variables. To address these existing challenges, we propose to use a separate encoder for each clinical variable for \textit{intra-variable} temporal dynamics, and then concatenate the outputs of the seperate encoders to learn the \textit{inter-variable} correlation and dependencies.

\subsection{Cross-Attention for Joint Learning from Multimodal Clinical Time Series}
To learn multimodal representations, we merged the embeddings of structured clinical time series and clinical notes using a validated cross-attention-based approach \citep{lu2019vilbert,husmann2022importance,liu2023attention,zhang2023improving}. Notes were encoded using a pretrained Longformer model\citep{li2022clinical}, which outperformed other pretrained text models such as ClinicalBERT.

%\begin{figure}[h]
%\centering
%\includegraphics[width=.7\textwidth]%{Figures/cross_attention.pdf}
%\caption{Overview of cross-attention fusion \citep{lu2019vilbert}. The inter-modality dependencies are captured by searching relevant information between modalities.}
%\label{fig:2}
%\end{figure}

%Figure~\ref{fig:2} shows an overview of cross-attention-based fusion. 
Consider the embeddings of clinical notes and structured EHR time series, denoted by $X_{note}$ and $X_{time}$, respectively. The core of this approach lies in generating enriched feature sequence by searching relevant information between modalities. For example, $W_{time \rightarrow note} = softmax(\frac{Q_{note}K_{time}^{T}}{\sqrt{d_{k}}})$ represent a scoring matrix, whose $(i, j)$-th element measures the attention given by the information from the $i$-th time step from modality $X_{note}$ and the $j$-th time step from modality $X_{time}$.

\section{Experiments}
In this section, we describe the evaluation of our approach on the benchmark task of predicting multiple in-hospital complications of major inpatient surgery using a real-world EHR dataset. Some technical details have been omitted for brevity and can be found in the Appendix.
\subsection{Dataset}
Our data consists of complete EHR records for all major inpatient surgeries occurring at three hospitals split among three hospitals (UF Health Gainseville, UF Health Jacksonville, and UF Health North Jacksonville) between 2014 and 2019. The combined cohort consisted of 113,953 adult patients who underwent 124,777 inpatient surgeries.

For each inpatient surgery, our dataset consists of: (1) the patient's demographic and admission information, such as age, sex, and body mass index; (2) 14 intraoperative time series consisting of vital signs such as blood pressure, heart rate, and body temperature; (3) all preoperative and intraopertive clinical notes for an encounter, such as History and Physical (H\&P notes) and operative reports; and (4) 9 binary labels indicating the occurrence of 9 postoperative complications.

\subsection{Benchmark Multitask Classification}
The goal is to predict the onset of nine postoperative complications following major inpatient surgery: prolonged ($>$ 48 hours) intensive care unit (ICU) stay, acute kidney injury (AKI), prolonged mechanical ventilation (MV), wound complications, neurological complications, sepsis, cardiovascular complications, venous thromboembolism (VTE), and in-hospital mortality. Models are trained on data available in the EHR up to the recorded timestamp of surgery end.

\subsection{Models}
Our embedding and tokenization scheme was designed for use with clinical transformers, and our primary classification model is a Longformer with sliding window attention. We compared the model trained with our dynamic embedding and tokenization scheme with several widely adopted baselines:

\textbf{Tokenized gated recurrent units (GRUs) with attention}: the tokenized sequential data was provided as input to a multi-layer GRU, followed by a self-attention layer \citep{tan2020data,shi2021deep}.

\textbf{Tokenized XGBoost}: an XGBoost model was trained on tokenized sequential data \citep{wang2020utilizing,liu2022dynamic}.

\textbf{BHERT}\citep{li2020behrt}, a widely used baseline transformer model for EHR data. In this baseline model, we removed variable-specific encoders and used one single embedding layer to encode sequential data. We also removed the relative positional embedding and time embedding, as they were not included in BHERT tokenization scheme.

\textbf{Hi-BEHRT} \citep{li2022hi}, a hierarchical transformer-based model for EHR data, which  employs a sliding window to partition the long sequence into smaller segments. Within each segment, a transformer was utilized as a local feature extractor to capture temporal dynamics.

\textbf{Self-supervised Transformer for Time-Series (STraTS)} \citep{tipirneni2022self}. STraTS uses a unique transformer to encode each variable with transformers and then uses a self-attention layer to generate the time-series embedding.

\section{Results and Discussion}
\subsection{Time Series Modeling}

\begin{table}[!htbp]
\centering
\caption{Model performance learning from clinical time series. Comparing the AUROC scores of different approaches for predicting nine postoperative outcomes.}
\label{tab:1}
\begin{adjustbox}{width = 1\textwidth}
\begin{tabular}{@{}ccccccccccc@{}}
\toprule
Task                & Mean  & ICU   & AKI   & MV    & Mortality & Wound & Neurological & Sepsis & Cardiovascular & VTE   \\ \midrule
GRU + Attention     & 0.771$\pm 0.04$  & 0.857 & 0.718 & 0.783 & 0.816 & 0.712 & 0.753 & 0.791 & 0.762 & 0.747 \\
XGBoost             & 0.765$\pm 0.03$ & 0.851 & 0.716 & 0.771 & 0.815 & 0.709 & 0.748 & 0.788 & 0.760 & 0.727 \\
Transformer (BEHRT) & 0.749$\pm 0.01$ & 0.843 & 0.701 & 0.765 & 0.800 & 0.701 & 0.725 & 0.770 & 0.748 & 0.699      \\
Hi-BEHRT             & 0.781$\pm 0.03$ & 0.863 & 0.730 & 0.789 & 0.835 & 0.721 & 0.769        & 0.801 & 0.780          & 0.769 \\
STraTS             & 0.797$\pm 0.04$ & 0.881 & 0.742 & 0.803 & \textbf{0.857} & 0.734 & 0.797        & \textbf{0.813} & 0.791          & 0.772 \\
Transformer (Ours)  & \textbf{0.801$\pm 0.02$} & \textbf{0.883} & \textbf{0.749} & \textbf{0.810} & 0.853 & \textbf{0.739} & \textbf{0.800} & 0.811  & \textbf{0.797} & \textbf{0.774} \\ \bottomrule
\end{tabular}
\end{adjustbox}
\end{table}

Table~\ref{tab:1} compares the area under the receiver operating characteristic curve (AUROC) for our benchmark multitask classification task. As shown in the table, our dynamic tokenization scheme-based transformer model outperform all baseline models with the highest mean AUROC of 0.801. We experimented with different types of variable-specific encoders (1-D convolutional layers \citep{kiranyaz20211d} and transformer layers \citep{tipirneni2022self}) and found similar results (For 1-D CNN, transformer, and linear, the mean AUROC score was 0.796, 0.800, and 0.798, respectively). STraTS \citep{tipirneni2022self} slightly underperformed our approach, suggesting the utility of the added relative positional embeddings used by our approach. With the same tokenized sequence, GRU + Attention (mean AUROC: 0.771) outperformed transformer models with traditional tokenization scheme (mean AUROC: 0.749), indicating the advantages offered by our embedding and tokenization framework for other non-transformer modeling approaches.

\subsection{Multimodal Fusion}

\begin{table}[!htbp]
\centering
\caption{Model performance using multimodal fusion of structured time series and free-text clinical notes. Shown are the AUROC of different approaches for predicting nine postoperative outcomes.}
\label{tab:2}
\begin{adjustbox}{width = 1\textwidth}
\begin{tabular}{@{}ccccccccccc@{}}
\toprule
Task & Mean & ICU & AKI & MV & Mortality & Wound & Neurological & Sepsis & Cardiovascular & VTE \\ \midrule
Time Series Only & 0.801$\pm 0.02$ & 0.883 & 0.749 & 0.810 & 0.853 & 0.739 & 0.800 & 0.811 & 0.797 & 0.774 \\
Clinical Notes Only & 0.821$\pm 0.02$ & 0.868 & 0.756 & 0.815 & 0.883 & 0.758 & 0.836 & 0.869 & 0.796 & 0.823 \\
Late Weighted Fusion & 0.813$\pm 0.03$ & 0.882 & 0.748 & 0.807 & 0.850 & 0.752 & 0.815 & 0.809 & 0.797 & 0.778 \\
Crossmodal Fusion + Concat\ref{fig:3} & 0.822$\pm 0.01$ & 0.866 & 0.755 & 0.816 & 0.881 & 0.754 & 0.831 & 0.867 & 0.797 & 0.831 \\
Concat + Crossmodal Fusion\ref{fig:4} & \textbf{0.845$\pm 0.03$} & \textbf{0.908} & \textbf{0.781} & \textbf{0.845} & \textbf{0.905} & \textbf{0.780} & \textbf{0.855} & \textbf{0.882} & \textbf{0.823} & \textbf{0.838} \\ \bottomrule
\end{tabular}
\end{adjustbox}
\end{table}

Table~\ref{tab:2} compares the AUROC of different models for multimodal learning of structured time series and free-text clinical notes. All of the multimodal models (time series + clinical notes) outperformed unimodal models utilizing either time series or clinical notes, aligning with conclusions from prior work\citep{husmann2022importance,lyu2022multimodal}. Concat + Crossmodal Fusion (shown in Appendix, Figure~\ref{fig:3}) performed the best, establishing a state-of-the-art mean AUROC of 0.845 for this task. Multimodal models trained without cross-attention resulted in less accurate predictions, suggesting this cross-attention-based fusion approach can effectively learn joint multimodal representations of time series data and clinical notes.

\section{Conclusion}
In this work, we introduced a dynamic embedding and tokenization scheme to adapt to the unique temporal challenges found in multimodal clinical time series. Experiments with real-world EHR databases on a benchmark clinical prediction task highlighted its advantages. Our work makes several contributions to the ongoing research of clinical time series modeling, as well as exploring innovative approaches to incorporate diverse health data sources.

\appendix
\section{Appendix}
Below is the appendex section, including multimodal architectures, dataset statistics, data preprocessing details, experiment details, and related work.

\subsection{Multimodal Architectures}
\begin{figure}[b]
\centering
\includegraphics[width=.7\textwidth]{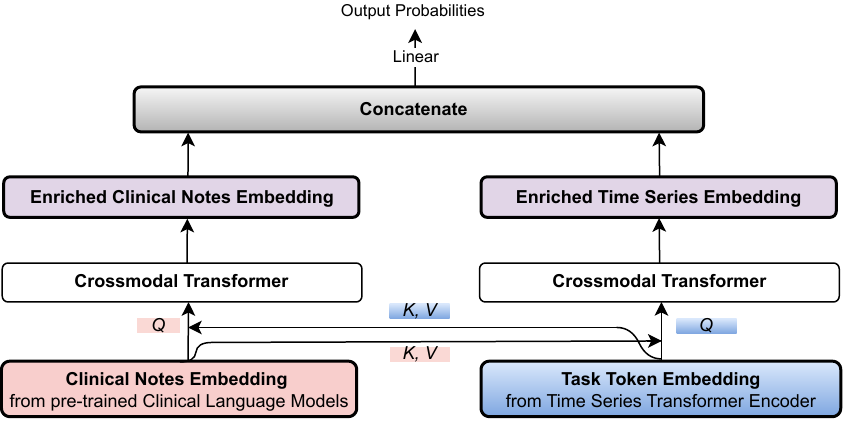}
\caption{Overview of Crossmodal Fusion + Concat.}
\label{fig:3}
\end{figure}

\begin{figure}[]
\centering
\includegraphics[width=.7\textwidth]{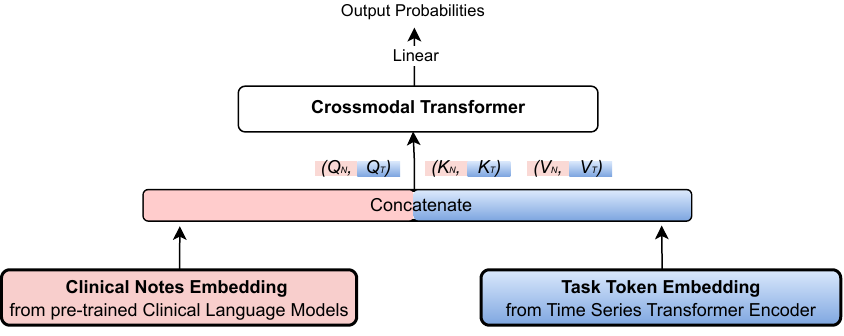}
\caption{Overview of Concat + Crossmodal Fusion.}
\label{fig:4}
\end{figure}

Please see Figure~\ref{fig:3} and Figure~\ref{fig:4} for two cross-attention-based fusion architectures.

\subsection{Dataset Statistics}
For each patient who underwent surgeries, we extracted following features:
\begin{enumerate}
    \item 9 preoperative demographic and admission information from 113,953 patients, including age (Mean 51 y, Min 18 y, Max 106 y), sex (48\% male, 52\% female), language, ethnicity, race, smoking status, zip code, and body mass index.

    \item 14 intraoperative temporal vital signs, including systolic blood pressure, diastolic blood pressure, mean arterial pressure, heart rate, respiratory rate, oxygen flow rate, fraction of inspired oxygen (FIO2), oxygen saturation (SPO2), end-tidal carbon dioxide (ETCO2), minimum alveolar concentration (MAC), positive end-expiratory pressure (PEEP), peak inspiratory pressure (PIP), tidal volume, and body temperature.

    \item 173 types of all preoperative and intraopertive clinical notes for an encounter, such as History and Physical (H\&P notes) and operative reports.

    \item 9 clinical outcomes, the incidence of complications include 23.29\% ICU stay (for 48 h or more), 13.09\% acute kidney injury, 8.64\% prolonged mechanical ventilation, 2.00\% in-hospital mortality, 13.48\% wound complications, 15.09\% neurological complications, 8.20\% sepsis, 12.18\% cardiovascular complications, and 4.51\% venous thromboembolism.
\end{enumerate}

\subsection{Data Preprocessing}
\textbf{Demographic and Admission Information}
For categorical data, we converted each to one-hot vectors, and concatenated with remaining numerical values. Missing static features was imputed with cohort medians.

\textbf{Time Series Data}
For 14 intraoperative time series data, their variable names were converted to unique integer identifiers; the measured values for each variable was were normalized to zero mean and unit variance based on the values from the training set; their measurement time, in the format of ``month/day/year hour:min:sec'', were first converted to unix timestamps and then also normalized similarly. For absolute positional indices, we assign one integer positional index for each token yet not enforcing the restriction that positional indices are unique and if different variables were measured at the same time. For relative positional embeddings, we generated the relative positional representation based on the GitHub code\footnote{\url{https://github.com/microsoft/MPNet/blob/master/pretraining/fairseq/modules/rel_multihead_attention.py}} for the original paper \citep{shaw2018self}.

\textbf{Clinical Notes} In the preprocessing phase, we merged all types of notes per surgery, converted the text to lowercase, and removed special characters and de-identification placeholders. Subsequently, we generated embeddings by first tokenizing the whole text using the clinically pretrained tokenizer. The tokens were then chunked to fit the pretrained clinical LLM, and the last hidden layer output for the CLS token was extracted as the embedding for each chunk. The final representation for each surgery was obtained by calculating the average of all these embeddings. We fixated on the Clinical Longformer \footnote{\url{https://huggingface.co/yikuan8/Clinical-Longformer}} for generating the embeddings due to its superior performance in classifying with clinical notes, following extensive testing with various models from Huggingface including BioBERT \footnote{\url{https://huggingface.co/dmis-lab/biobert-base-cased-v1.2}}, BiomedBERT\footnote{\url{https://huggingface.co/allenai/biomed_roberta_base}}, ClinicalBERT \footnote{\url{https://huggingface.co/emilyalsentzer/Bio_ClinicalBERT}}, and Clinical Longformer\footnote{\url{https://huggingface.co/yikuan8/Clinical-Longformer}}.

\subsection{Experiment Details}
\textbf{Multi-task Training}
Our model was trained with the multi-task fashion for predicting 9 postoperative outcomes. To do this, we expanded the notion of ``[CLS]'' token for text classification and prepended 9 global tokens to our tokenized sequences, one for each of our postoperative outcomes, so that self-attentions were computed among all sequence elements for each clinical outcome token.

\textbf{Longformer for processing long sequences} Vanilla transformer models are largely limited by their capacity in processing long sequences (\textit{max\_sequence} = 512) because of its quadratic complexity with respect to the sequence length. Therefore, in this paper we used Longformer \citep{beltagy2020longformer} for time series modeling transformers. Longformer architecture introduces sliding window attention that allows the model to attend to only a subset of tokens within a window, reducing the overall computational complexity. This design makes Longformer well-suited for processing longer sequences (\textit{max\_sequence} = 4096).

\textbf{Experimental Setup}
We trained the models on two NVIDIA A100-SXM4-80GB GPUs for 30 epochs to leverage hardware acceleration. We used a batch size of 32 per GPU for the best performing model.

\textbf{Hyperparameters}
We used the following hyperparameters for optimization and regularization, Adam optimizer with a learning rate of 1e-4, dropout of 0.2, and weight decay of 1e-4. For the transformer models, including the Longformer, we limited the models to only 1 attention head and 1 layer per head, as this configuration produced the best results.

\textbf{Loss Function}
We chose to use a Binary Cross-Entropy with Logits (BCEWithLogitsLoss) with the parameter \textit{pos\_weight} set to the weight of the positive class for each prediction outcome, given the unbalanced nature of our dataset. This allows models to be more sensitive to minority class by increasing the cost of misclassification of it.

\subsection{Related Work}
\textbf{Transformers for EHRs} Recent transformers for EHRs focused on adapting vanilla transformer architecture for tokenizing irregular sampling and learning temporal dependencies in long sequences. For example, for sparsity and irregular sampling, \citet{zhang2021graph} used graph neural network to embed irregularly sampled and multivariate time series, capturing sensor dynamics from observational data. \cite{tipirneni2022self} treated time-series as sets of observation triplets and uses transformers to encode continuous time and variable values, eliminating the need for discretization. For processing long sequences, \citet{li2022hi} employed a sliding window to partition the complete medical history into smaller segments. Within each segment, a Transformer was utilized as a local feature extractor to capture temporal interactions. Motivated by this line of research, we introduced our dynamic embedding approache in this paper, aiming to address temporal EHR challenges by learning time-aware representations patient time series.

\textbf{Multimodal Representation Learning for Health}
Combining diverse sources of data sources in medical domain is promising for more comprehensive understanding of patients' health conditions \citep{raghupathi2014big}, more accurate health outcome predictions \citep{shickel2023dawn}, and building next-generation foundational medical models for generative AI \citep{moor2023foundation}. The core of this research effort is multimodal representation learning where all the modalities are projected to a common space while preserving information from the given modalities \citep{liang2022high}. Traditional data fusion methods, such as early and late fusion, are insufficient to learn the correlations and dependencies among different modalities \citep{ma2022detecting}. Recently, transformer-based architecture, thanks to its superior ability to capture cross-modal interactions by self-attention and its variants \citep{xu2023multimodal}, has achieved great success in various multimodal machine learning tasks in different domains, such as multimodal action recognition \citep{wang2020transmodality}, image segmentation \citep{xiao2023transformers}, and affect detection \citep{ma2023noisy}.

%\subsubsection*{Author Contributions}
%author contributions

\subsubsection*{Acknowledgments}
We would like to thank the NVIDIA Corporation for providing computational resources used for this research. This research was also supported by National Institute of Health (NIH) through grant R01 GM110240. Any opinions, findings, conclusions, or recommendations expressed in this research are those of the authors, and do not necessarily represent the official views, opinions, or policy of NIH.
\bibliography{iclr2024_conference}
\bibliographystyle{iclr2024_conference}

\end{document}